\documentclass[final]{l4dc2026}

\title[Learning to Act through Contact]{Learning to Act Through Contact: \\ A Unified View of Multi-Task Robot Learning}
\usepackage{times}

\coltauthor{\Name{Shafeef Omar} \Email{shafeef.omar@tum.de}\\
 \Name{Majid Khadiv} \Email{majid.khadiv@tum.de}\\
 \addr Munich Institute of Robotics and Machine Intelligence, Germany \\ Technical University of Munich, Germany}

\begin{document}

\maketitle

\begin{abstract}%
 We present a unified framework for \emph{multi-task} locomotion and manipulation policy learning grounded in a contact-explicit representation. Instead of designing different policies for different tasks, our approach unifies the definition of a task through a sequence of contact goals--desired contact positions, timings, and active end-effectors. This enables leveraging the shared structure across diverse contact-rich tasks, leading to a single policy that can perform a wide range of tasks. In particular, we train a goal-conditioned reinforcement learning (RL) policy to realise given contact plans. We validate our framework on multiple robotic embodiments and tasks: a quadruped performing multiple gaits, a humanoid performing multiple biped and quadrupedal gaits, and a humanoid executing different bimanual object manipulation tasks. Each of these scenarios is controlled by a single policy trained to execute different tasks grounded in contacts, demonstrating versatile and robust behaviours across morphologically distinct systems. Our results show that explicit contact reasoning significantly improves generalisation to unseen scenarios, positioning contact-explicit policy learning as a promising foundation for scalable loco-manipulation. Video available at: \url{https://youtu.be/idHx67oHHU0?si=qZJ7C0ujemXNWgA5}
\end{abstract}


\section{Introduction}

Advances in reinforcement learning (RL) have enabled robots to master complex motor skills, from agile quadruped locomotion \cite{hoeller2023anymalparkourlearningagile, cheng2023parkour} to dexterous object manipulation \cite{openai2019solvingrubikscuberobot, singh2025dextrahrgbvisuomotorpoliciesgrasp, yin2025dexteritygenfoundationcontrollerunprecedented}. Yet, prevailing RL policies are often trained with task-specific objectives, making them difficult to transfer to unseen scenarios without retraining from scratch. For example, perceptive locomotion policies are typically tasked to train on velocity and/or position commands that work well on various rough terrain scenarios they have been trained on \cite{rudin2022learningwalkminutesusing, miki2022perceptivequadrupedallocomotion}. Nonetheless, they cannot be directly transferred to environments with sparser footholds and riskier terrains \cite{zhang2024learningagilelocomotionrisky}, such as stepping stones, even though the required motions are similar to those on which it has been trained on. Similarly, in object manipulation, tasks like lifting an object from a table share similar motor skills with more complex tasks, such as stacking objects. However, the traditional approach of training policies on specific tasks makes it difficult to generalize across different types of physical interactions. Moreover, training a robot on a new task from scratch is not feasible each time it encounters one. This motivates seeking a more fundamental abstraction: one that also unifies locomotion and manipulation. In this light, we propose using \textit{contacts} as a common representation to enable better generalization and adaptability across various physical interaction tasks and embodiments.

    \textbf{Why Contacts?} Contacts govern nearly all loco-manipulative behaviors. Locomotion requires coordinated foot placements with the ground, and manipulation relies on purposeful hand-object interactions—both inherently contact-driven. Despite their centrality, contacts are often treated as incidental in reinforcement learning (RL) frameworks, emerging implicitly as a byproduct of motion optimization. This abstraction leads to policies with limited generalization across tasks that share underlying motor principles. In contrast, humans intuitively decompose complex behaviors into contact-explicit subgoals: a climber plans handholds and footholds before ascending, and a parkour athlete sequences hand and foot placements relative to environmental features. These skills transfer seamlessly across structurally similar tasks.

    While recent works have explored goal representations such as 3D position targets~\cite{sferrazza2024humanoidbench} or motion references~\cite{he2024hover, yin2025dexteritygenfoundationcontrollerunprecedented}, they often overlook contact as a fundamental primitive. As a result, such approaches may struggle with tasks where contact timing and placement are crucial. Recent works have shown the benefits of explicitly incorporating contact—either in rewards or task representations~\cite{zhang2024wococo,ciebielski2024contact, lin2025sim}—leading to better generalization and performance.


    \begin{figure*}[t]
        \centering
        \includegraphics[width=\linewidth]{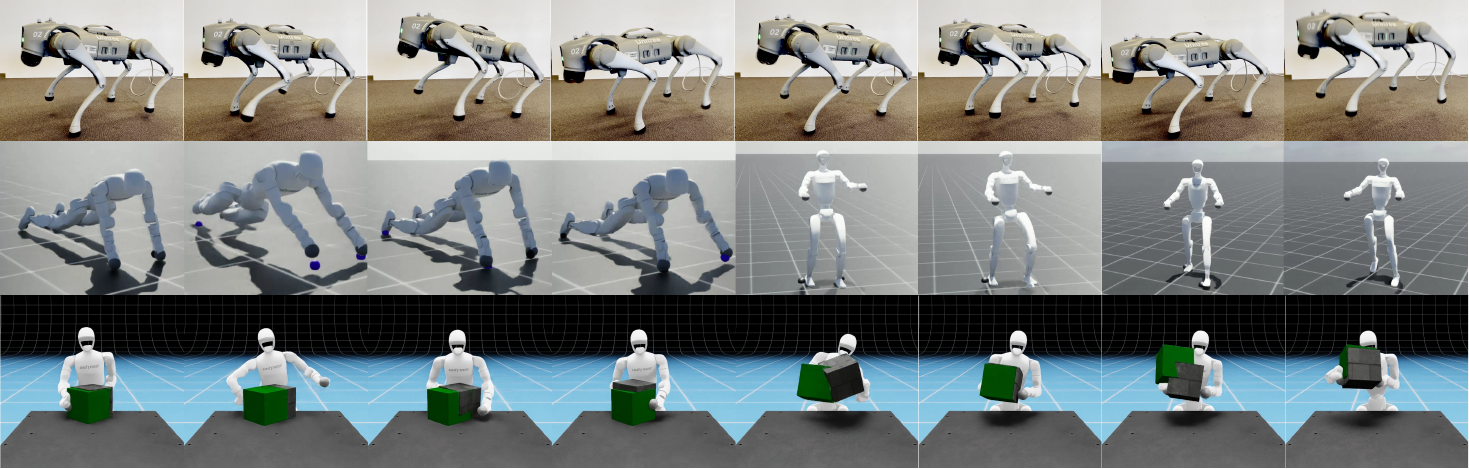}
        \caption{\centering Snapshots of our contact-explicit framework in action. (Row 1): the quadruped demonstrates diverse gaits; (Row 2): the humanoid demonstrates locomotion using hand-assisted gaits such as quadrupedal jump, pace and bipedal gaits such as walk, jump; (Row 3): the humanoid carries out different bimanual manipulation tasks, where the green box represents the target pose.}
        \label{fig:contact-firstpage}
    \end{figure*}
    
    Our work builds on this contact-explicit paradigm, proposing contact goals as a unified task representation for locomotion and contact-rich manipulation, enabling a single policy to produce diverse physical behaviors. We decompose tasks into contact goals—defined by \textit{target locations, timings}, and \textit{active end-effectors}—together with object pose targets for manipulation. Given a high-level planner that generates sequences of these goals, a goal-conditioned RL policy learns to achieve them via joint torques across various contact modes. This bypasses task-specific reward design in the policy, instead treating contacts as fundamental physical primitives which robots interact with their environment.

    We validate our framework with three demonstrations, each with a single policy: (1) on a quadruped robot to execute multiple gaits (trot, pace, bound, jump, and crawl), (2) on a humanoid robot performing multiple biped and quadruped gaits, and (3) on a humanoid robot to perform bi-manual object manipulation such as object reorientation on a table and object pose tracking while being lifted. Our results demonstrate that using a contact-explicit representation, we can generate multiple skills with a single policy and leverage the shared information between tasks to improve generalization beyond the training distribution.


Our contributions can be summarised as follows:
\begin{enumerate}
    \item We propose a goal-conditioned RL framework that learns to achieve given contact goals with the world. Using this framework, we train a single policy capable of performing multiple tasks.
    \item We provide empirical validation on morphologically distinct robots and various tasks, showing that explicit contact reasoning enables dynamic and robust behaviors across diverse scenarios, while generalizing to unseen tasks. 
\end{enumerate}
\section{Related Work}
\label{sec:related_work}

While locomotion and contact-rich manipulation are similarly realized through intermittent contacts with the environment, RL-based methods typically use different specialized rewards and task representations for each problem. \\ \textbf{Locomotion. } Early works represented predefined gaits (walking) with step location and robot heading as input~\cite{peng2017deeploco}. More recent works avoid predefining gaits, and define the desired behavior (goal) through a desired average velocity~\cite{hwangbo2019learning, rudin2022learningwalkminutesusing} or reaching a desired position in the world \cite{rudin2022advancedskillslearninglocomotion, hoeller2023anymalparkourlearningagile, cheng2023parkour}. As this goal representation does not distinguish between different gaits, it is not suitable for multi-gait policy generation and its performance is highly constrained to its training distribution. To enable one policy for multiple gaits, recent approaches used either a notion of gait phase as input to the policy~\cite{margolis2022walktheseways,bellegarda2024allgaits} or task-specific desired reference motion~\cite{tan2018sim,li2024reinforcement, zargarbashi2024robotkeyframinglearninglocomotionhighlevel, sleiman2024guided}. The former representation is specific to locomotion and cyclic gaits, while the latter requires another module to generate desired trajectories for every behavior, which could become expensive. Unlike these approaches, our findings reveal that we do not need to use any parameterisation or dense reference trajectories to represent the different gaits, but rather intuitive contact goals can directly achieve them.

\textbf{Manipulation/Loco-Manipulation. } In manipulation, the desired behavior is usually specified through the desired object goal~\cite{openai2019solvingrubikscuberobot, lin2023bi} or task-specific goals such as for grasping~\cite{lum2024dextrahgpixelstoactiondexterousarmhand, singh2025dextrahrgbvisuomotorpoliciesgrasp}. While successful in learning single tasks, such a representation fails to work in most multi-task and the few-shot learning settings~\cite{chen2022towards}.  In loco-manipulation settings, most works simply concatenate separate locomotion and manipulation goals \cite{pan2025roboduet, fu2022deepwholebodycontrollearning} or define and train different tasks separately \cite{dao2024sim,liu2024visual,qiu2024wildlma}. However, such an approach does not enable leveraging the shared structure between locomotion and manipulation through contact. \cite{sferrazza2024humanoidbench} trained whole-body controllers using 3D position targets for the robot's hands and then trained a high-level planner using these to perform loco-manipulation. They also released a suite of complex tasks as a benchmark for robot loco-manipulation. However, their low-level reaching policy ignores contacts, which are crucial for loco-manipulation and, consequently, fails in many tasks on their benchmark. 

\textbf{Contacts. }Recent studies have shown that including contact information in the reward design and task representation can improve multi-task learning~\cite{zhang2024wococo,ciebielski2024contact, lin2025sim}. In particular, \cite{ciebielski2024contact} showed that a contact-centric representation for multi-gait locomotion learning improves the generalization capability of the gaits when compared to other representations. However, they only showed locomotion results in a behavioral cloning setting. \cite{zhang2024wococo} used the contact information in the reward design for various locomotion and loco-manipulation tasks. They proposed sparse contact-based rewards that are then combined with task-specific rewards to enable complex motions such as humanoid parkour and loco-manipulation. Compared to \cite{zhang2024wococo}, which learns different policies for different tasks, we show that training one multi-skill policy outperforms the generalization capabilities of the policy to unseen tasks. Furthermore, different from \cite{zhang2024wococo}, we present a denser reward for contact that facilitates the training procedure and qualitatively produces smoother motions. Closest to our work is a recent study that also uses contact and object pose goals to train an RL policy only to perform bimanual dextrous manipulation \cite{lin2025sim}. We show that our proposed contact-conditioned policy generalizes better than the (sub)task-conditioned policies in \cite{lin2025sim} for object manipulation. Furthermore, we show that contact-conditioned policies are general enough to be applied to the locomotion setting as well. 

\section{Method}
\label{sec:method}
\subsection{Overview}

\begin{figure}[t]
	\centering
	\includegraphics[width=0.65\linewidth]{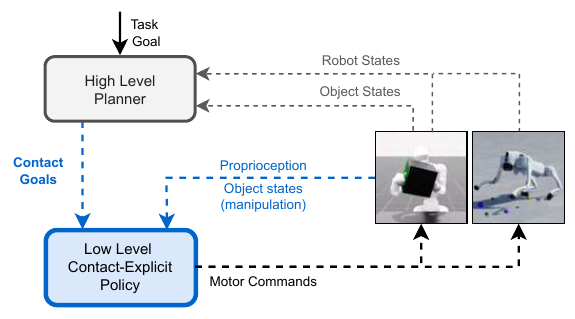} 
	\caption{ Overview of our contact-explicit framework. A high-level planner generates the contact goals (and object pose targets for manipulation), that is provided as immediate goals for the goal-conditioned RL policy to accomplish. }
	\label{fig:framework}
	\vspace{-1.0em}
\end{figure}


    At the core, we propose a contact-explicit representation that is used to train policies for multi-gait locomotion on a quadruped/humanoid and a multi-task bimanual manipulation on a humanoid robot. In particular, we train a goal-conditioned RL policy to track contact goals, provided by a planner as shown in Fig.~\ref{fig:framework}. The \textbf{contact goals} for an end-effector $e$, $g_e^{\text{con}} = \{p_e^{\text{con}}, S_e^{\text{con}}, \mathcal{I}_e^{\text{con}}\}$, correspond to the 3D location of contact, the contact duration, and a binary indicator to be in contact, respectively. In the case of manipulation, $g_e$ additionally comprises the object's goal pose $\{p_{obj}, \theta_{obj}\}$. A new set of contact goals is chosen when the contact duration expires and it achieves the desired contact. As such, by composing several different contact goals, we can perform various long-horizon tasks. In this work, we have prespecified the contact goals required to achieve the various tasks. However, our method can be integrated with more sophisticated learned contact planners~\cite{omar23humanoids, dhedin2024diffusion}, or even contact goals extracted from images/videos~\cite{taouil2025physicallyconsistenthumanoidlocomanipulation}. 

    We formalize the problem of finding a multi-task policy \(\pi(a_t|s_t, g_t)\) as a goal-conditioned RL problem  which is formulated as a Markov Decision Process (MDP) $\mathcal{M} = \langle \mathcal{S}, \mathcal{A}, \mathcal{T}, \mathcal{R}, \gamma, \mathcal{G} \rangle$. This MDP is defined by states $s_t \in \mathcal{S}$, actions $a_t \in \mathcal{A}$, transtion probability $\mathcal{T}$, a reward $r_t \in \mathcal{R}$, a discount factor \(\gamma\) and goal $g_t \in \mathcal{G}$. The reward \(r_t\) is calculated to achieve different contact modes by following the immediate contact goals. The policy aims to maximise the expected return for achieving the contact goal \( g_t \):
    \[\max \mathbb{E}_{\pi} \left[ \sum_{t} \gamma^t r_t(s_t, a_t, g_t) \right]
\]
    



    \subsection{Learning to Act Through Contact}
    \label{subsec:contact-phases}

\begin{figure}
        \centering
        \includegraphics[width=0.4\linewidth]{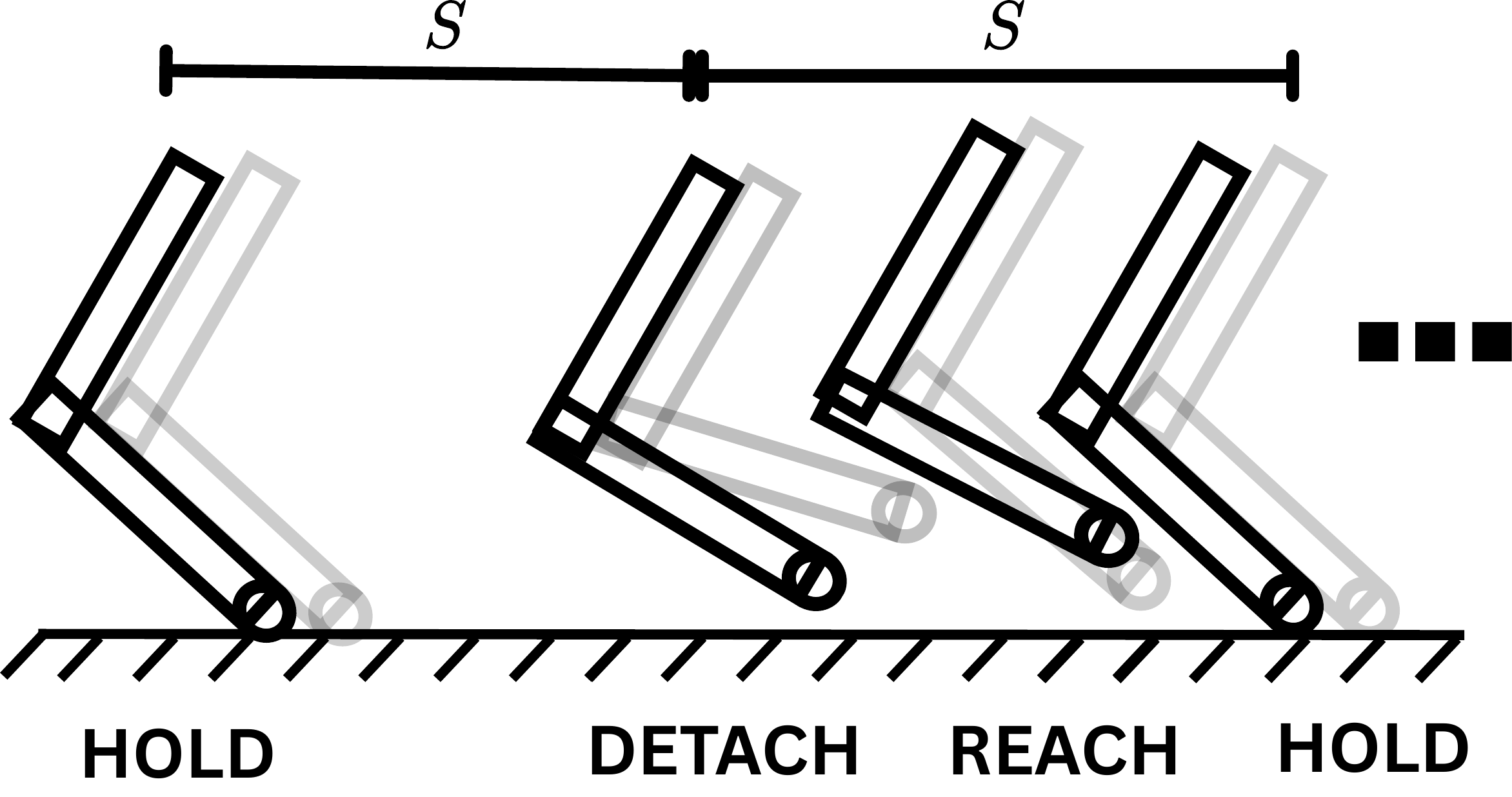} 
        \caption{Simple illustration of a robot's end effector during different phases of contact, for a fixed command duration $S$. During the detach phase, the end effector is detached from a contact and is free to move. During the reach phase, the end effector is guided towards the desired contact location. During the hold phase, it maintains the contact.}
        \label{fig:contact-phases}
    \end{figure}

    \textbf{Contact Phases.} We consider three phases for contact that allow us to make or break contact with the environment using any end-effector in a controlled manner, as illustrated in Fig.~\ref{fig:contact-phases}: reach (R), hold (H) and detach (D). During the reach phase of an end-effector, the robot must guide it to a desired contact location provided by the high-level planner. During the hold phase of an end-effector, the robot must maintain its contact where it was guided to during the reach phase. And during the detach phase of an end-effector, the robot is free to move it as long as it does not engage in contact. Viewing contacts from this perspective, we can develop dense rewards that allow the robot to explore several contact modes by making and breaking contacts with the world.
    
    For an end-effector $e$ at time $t$, the contact goals from the high-level planner comprise the following information: contact locations with a short horizon of two contact switches, \(p^{\text{con}}_{t, e} = \left( \left[p^{\text{con}}_{t, e}\right]_1, \left[p^{\text{con}}_{t, e}\right]_2 \right)\), a binary indicator of contact for two contact switches, \
    $(\mathcal{I}^{\text{con}}_{t, e} = \left( \left[\mathcal{I}^{\text{con}}_{t, e}\right]_1, \left[\mathcal{I}^{\text{con}}_{t, e}\right]_2 \right)\) and the command duration $S$ of the current contact goal to be achieved. The contact phase of an end-effector is determined using the binary indicator, $\left[\mathcal{I}^{\text{con}}_{t, e}\right]_1$, and the time remaining to finish the contact command, $s$ (where $s$ is reset to the value of the newly sampled command duration when the previous one expires). If the remaining command duration is less than a threshold $\delta$ and the binary contact indicator is 0, we have the reach phase ($\left[\mathcal{I}^{\text{con}}_{t, e}\right]_1 = 0$ and $s < \delta$). If the binary contact indicator is 1, we have the hold phase ($\left[\mathcal{I}^{\text{con}}_{t, e}\right]_1 = 1$). We have the detach phase if the binary contact indicator is 0 and the remaining command duration is greater than the threshold $\delta$ ($\left[\mathcal{I}^{\text{con}}_{t, e}\right]_1 = 0 $ and $s > \delta$). The binary contact indicators of all the end-effectors, $\mathcal{I}^{\text{con}}_{t, e}$, are stacked to form the contact sequence, referred to in the paper.
    
    \label{sec:result}
    \begin{figure*}[t]
    	\centering
    	\includegraphics[width=0.8\linewidth]{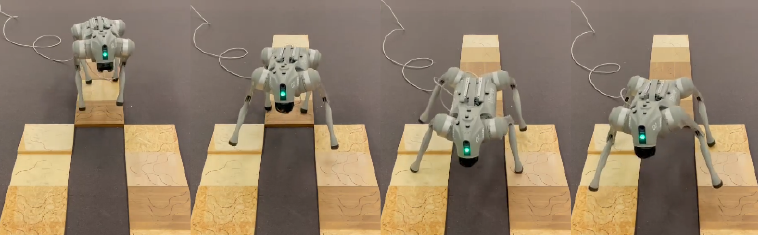} 
    	\caption{Quadruped robot crossing a gap with a bound gait by accurately adjusting the contact locations to remain on the wooden terrain.}
    	\label{fig:crossing_gap}
    \end{figure*}
    
    \textbf{Policy Observations and Rewards.} Apart from the proprioceptive inputs for the policy's observations, we provide explicit contact goals as described above, $p^{\text{con}}_{t, e}$, $\mathcal{I}^{\text{con}}_{t, e}$, and $s$, to achieve multiple tasks using the same shared structure. The following rewards are used to achieve the various contact phases described previously to perform multiple tasks:



\begin{equation}
	\resizebox{0.6\textwidth}{!}{$
		r_{t,e}^{\text{reach}} =
		\exp\left(-\frac{d\big(\left[p^{\text{con}}_{t, e}\right]_1, p^{\text{act}}_{t, e}\big)}{\sigma^2}\right)
		\cdot \mathbb{I}\left[\left[\mathcal{I}^{\text{con}}_{t,e}\right]_1 = 0 \land s \leq \delta \right]
		$}
\end{equation}

\begin{equation}
\resizebox{0.7\textwidth}{!}{$
r_{t,e}^{\text{hold}} = 
\left(1 + \alpha_{\text{hold}} \cdot \exp\left(-\frac{d\big(\left[p^{\text{con}}_{t, e}\right]_1, p^{\text{act}}_{t, e}\big)}{\sigma^2}\right)\right)
\cdot \mathbb{I}\left[\left[\mathcal{I}^{\text{con}}_{t,e}\right]_1 = \mathcal{I}^{\text{act}}_{t,e} = 1\right]
$}
\end{equation}

\begin{equation}
\resizebox{0.41\textwidth}{!}{$
r_{t,e}^{\text{detach}} = 
\mathbb{I}\left[\left[\mathcal{I}^{\text{con}}_{t,e}\right]_1 = \mathcal{I}^{\text{act}}_{t,e} = 0 \land s > \delta\right]
$}
\end{equation}

    where $d(\mathbf{a}, \mathbf{b})$ is the $L2$-norm between $\mathbf{a}$ and $\mathbf{b}$, 
    and $\mathbb{I}(\cdot)$ is an indicator function.
    The hold reward incentivises the robot to maintain contact, and it gets a higher reward for making contact at the desired location for $\alpha_{hold} > 0$. The detach reward is a scalar reward that incentivises the agent to not make any contact during this phase.

    For the case of manipulation, we additionally have a reward for tracking the object pose:

    \[
        r_{t, obj}^{pose} = \frac{c_{pos}}{\epsilon_{pos} + \Delta p_{t, obj}} + \frac{c_{rot}}{\epsilon_{rot} + \Delta \theta_{t, obj}}
    \]

    The total contact reward $r_t^{\text{con}}$ is given as:

    \[
        r_t^{\text{con}} = r_{t, obj}^{\text{pose}} + \sum_{e} \Big( r_{t, e}^{\text{reach}} + r_{t, e}^{\text{hold}} + r_{t, e}^{\text{detach}} \Big) 
    \]


\section{Experiments and Results}

    We evaluate our contact-explicit framework for performing multiple tasks on various robotic embodiments, such as a quadruped and a humanoid, and further conduct extensive experiments on them for locomotion and bimanual manipulation. Our evaluations are based mainly on the multi-tasking and representation capabilities of our contact-explicit approach.

    \textbf{Locomotion.} The quadruped is trained to perform multiple gaits, such as \textit{trot, pace, bound, jump and crawl}, as depicted in row 1 of Fig \ref{fig:contact-firstpage}.  The contact locations are sampled to move in all directions. First, stride lengths and stance widths for each pair of front and hind legs are sampled for each environment upon initialization from a uniform distribution $\mathcal{U}(0.0, 0.3)m$ and $\mathcal{U}(0.1, 0.3)m$, with a sampled heading direction $\mathcal{U}[-\pi, \pi] rad$ or a with a sampled yaw rate to follow curved paths $\mathcal{U}[-\pi, \pi] rad/s$. From these locations, we further sample additional offsets to allow more versatility for each leg, $\mathcal{U}(-0.15, 0.15)m$, both in lateral and longitudinal directions. We find that sampling these values once during initialisation leads to a better policy compared to sampling upon every reset. The command durations were sampled from a narrow uniform distribution of $[0.34, 0.36]$ seconds. We use extensive domain randomisation \cite{tan2018sim} to deploy our policy in the real world without additional fine-tuning. Fig \ref{fig:crossing_gap} shows our contact-explicit policy deployed to navigate a challenging gap crossing scenario while remaining on the wooden platform, highlighting accurate tracking. A similar strategy as described above was also used to train the humanoid robot for various biped (walk and jump) and quadruped (crawl, pace, jump) locomotion modes. 
    
    For both the embodiments, the policy observes proprioceptive states such as joint positions, joint velocities, and task observations such as the current and next contact sequence of all feet, the current and next contact locations of all feet in base frame, the command duration, relative distance of the robot's feet to its desired contact locations. Notably, we do not use any contact sensing on the robot’s feet since it did not make any difference in simulation performance. Apart from the rewards mentioned in \ref{subsec:contact-phases}, we additionally add penalties typically used in locomotion settings such as the base angular velocities, joint velocities, accelerations, torques, joint deviations, and action rate for smoother behaviours. We also update the goals if the robot's base when projected to the ground remains within a threshold of the desired contact locations and provide a bonus reward for discovering more goals.
    
    \textbf{Bimanual Manipulation.} Using our method, the humanoid is trained to perform the following bi-manual manipulation tasks as shown in row 3 of Fig.~\ref{fig:contact-firstpage}: 1) \textit{Repose}: The robot must continuously maintain contact on two surfaces of a box (cuboid) and must track several positions and orientations for the object, sampled from a uniform distribution. These object poses are in the air, hence the humanoid must learn to lift the object and not let it slip from its hand while tracking the various poses. 2) \textit{Reorient}: The robot must make and break contact with the box repeatedly to keep rotating it $45^{\circ}$ on the table each time it makes contact. The contact locations were predefined so that the correct surfaces could be chosen to rotate the object continuously. The command durations are sampled from a uniform distribution of $\mathcal{U}[1.0, 1.5] s$. 
    
    The policy observes proprioceptive robot states, object states, and task observations such as the current and next contact sequence of the robot's hands, the current and next contact location on the object's surface, the command duration, and the goal pose relative to the object's pose. Apart from the rewards mentioned in \ref{subsec:contact-phases}, we additionally provide penalties to penalise joint velocities, accelerations, torques, action rate and high impact forces. We only update the goals if the desired object pose is within a threshold of the actual object pose and provide a bonus reward for it.

    \textbf{Training. }For all the tasks, we use the Proximal Policy Optimization (PPO)  \cite{schulman2017proximalpolicyoptimizationalgorithms} algorithm 
    with a recurrent architecture (GRU) and entropy decay to train the policy in IsaacLab \cite{mittal2023orbit} with $8192$ parallel environments. We use PPO as our algorithm of choice since it is effective in learning low-level motion primitives, as also observed in \cite{yin2025dexteritygenfoundationcontrollerunprecedented}. In the future, we'd also like to explore off-policy RL algorithms to make use of goal relabelling \cite{andrychowicz2018hindsightexperiencereplay}. 
    
    \begin{figure*}[t]
	\centering
	\includegraphics[width=0.86\linewidth]{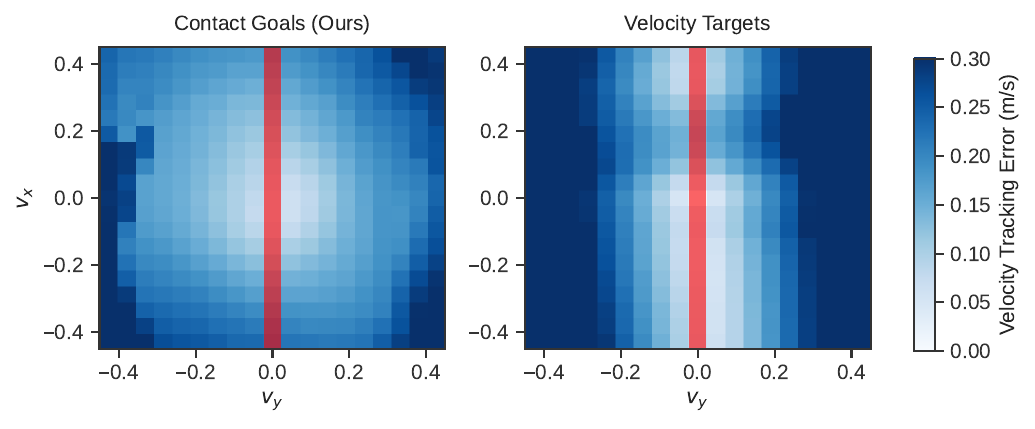}
	\caption{Comparison of velocity tracking error in all $x$-$y$ directions. Each cell in the grid is a combination of $x$ and $y$ velocity. The red line denotes the velocity combinations seen during training. The results were averaged over 500 episodes (each lasting 15 seconds of simulation time).}
	\label{fig:velocity-tracking-error}
\end{figure*}

	\begin{figure*}[t]
		\centering
		\includegraphics[width=0.8\linewidth]{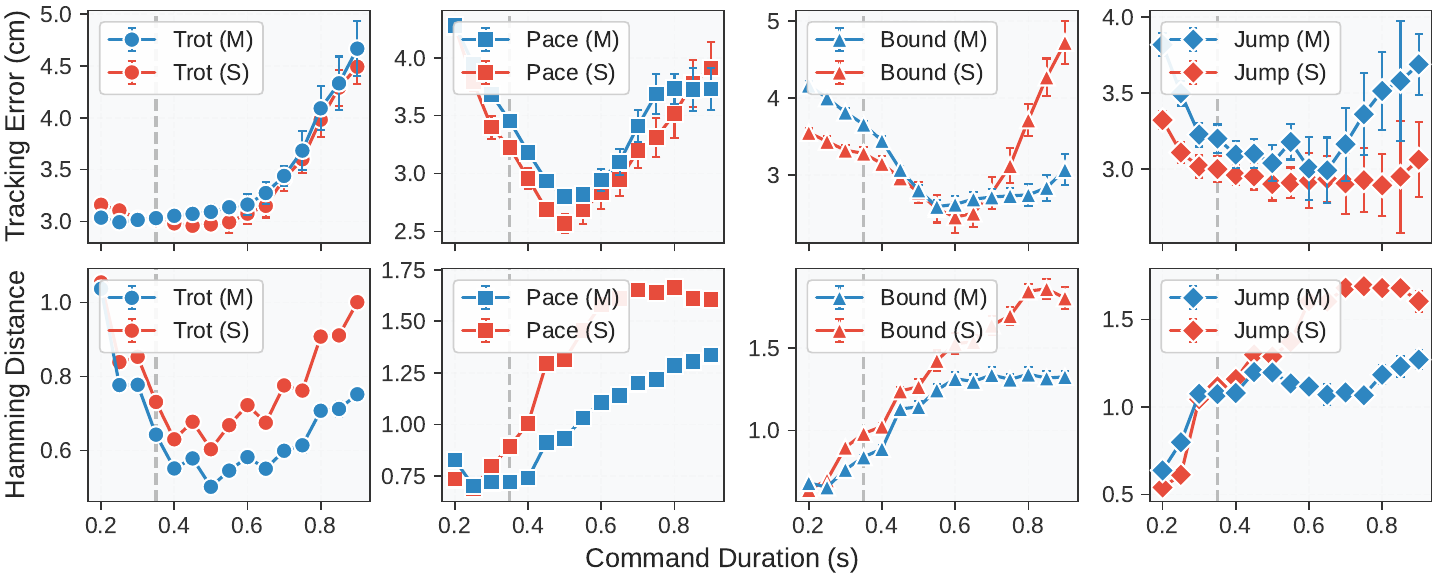}
		\caption{Comparison of contact location tracking ($L2$-norm) and contact plan deviation (Hamming distance) for the \textcolor{blue}{multi-gait policy} (denoted by \textcolor{blue}{M}) against \textcolor{red}{single-gait policies} (denoted by \textcolor{red}{S}) trained exclusively on one gait, evaluated over a broader range of command durations. Results are averaged over 1000 episodes, each lasting 15 seconds of simulation time. The dotted vertical grey line indicates the command duration seen during training.}
		\label{fig:multigait-vs-singlegait}
		
	\end{figure*}


    
    \textbf{Locomotion generalization to unseen velocities/directions.} To demonstrate that contact-explicit representations truly generalize better to out-of-distribution scenarios and has better representation capabilities, we compare two policies with different goal representations, contact-explicit (ours) trained to only perform trotting gait against velocity targets (typical in learning quadrupedal locomotion as in \cite{hwangbo2019learning, rudin2022learningwalkminutesusing} and known to converge to a trotting gait), each trained only to move forward/backwards up to a maximum speed of 0.65 $m/s$ (i.e., $v_x \in [-0.65, 0.65]$ $m/s$, $v_y = 0$). We evaluate the velocity tracking error of these two policies when commanded to move along all directions as shown in Fig.~\ref{fig:velocity-tracking-error}. 

    Although the policy trained with velocity targets has slightly less tracking error while extrapolating velocity targets along the trained direction, it becomes apparent that the contact-explicit policy can cover a much broader range of velocities, compared to the one trained with velocity targets. Especially in the case of lateral/sideways walking ($v_x = 0$), the velocity-conditioned policy hesitates to move. In contrast, the contact-explicit policy moves sideways, even though it hadn't received any reward for the lateral movements during training, but was solely trained to track the contact goals.

    \textbf{Multi-task versus single-task.} Contact-explicit task representation enables learning multiple gaits in a single policy, which helps leverage the shared structure between different gaits to interpolate between them (even though it has not seen those states during training). To test our claims, we compare our multi-gait quadrupedal locomotion policy against separate policies trained on single gaits, as shown in Fig.~\ref{fig:multigait-vs-singlegait}. The policies were trained with command durations sampled from a narrow uniform distribution of range $[0.34s, 0.36s]$ and evaluated over a broader range of $[0.2s, 0.9s]$. We use two metrics: contact location tracking error measured using $L2$-norm between the actual end-effector locations and the planned contact locations while making contact, and the contact plan deviation measured using the Hamming distance between the desired contact plan and the actual contact status of the end-effectors.
    
    From Fig.~\ref{fig:multigait-vs-singlegait}, we observe that the multi-gait (M) policy has the lowest contact plan deviation across all the evaluated command durations and for all gaits. We hypothesize that this is mainly due to our contact-explicit representation that enables learning multiple gaits in a single policy. We also observe that generally, the tracking error of both the single-gait and multi-gait policies are similar, i.e. within 2 cm of difference.
    \begin{figure*}[t]
		\centering
		\includegraphics[width=0.8\linewidth]{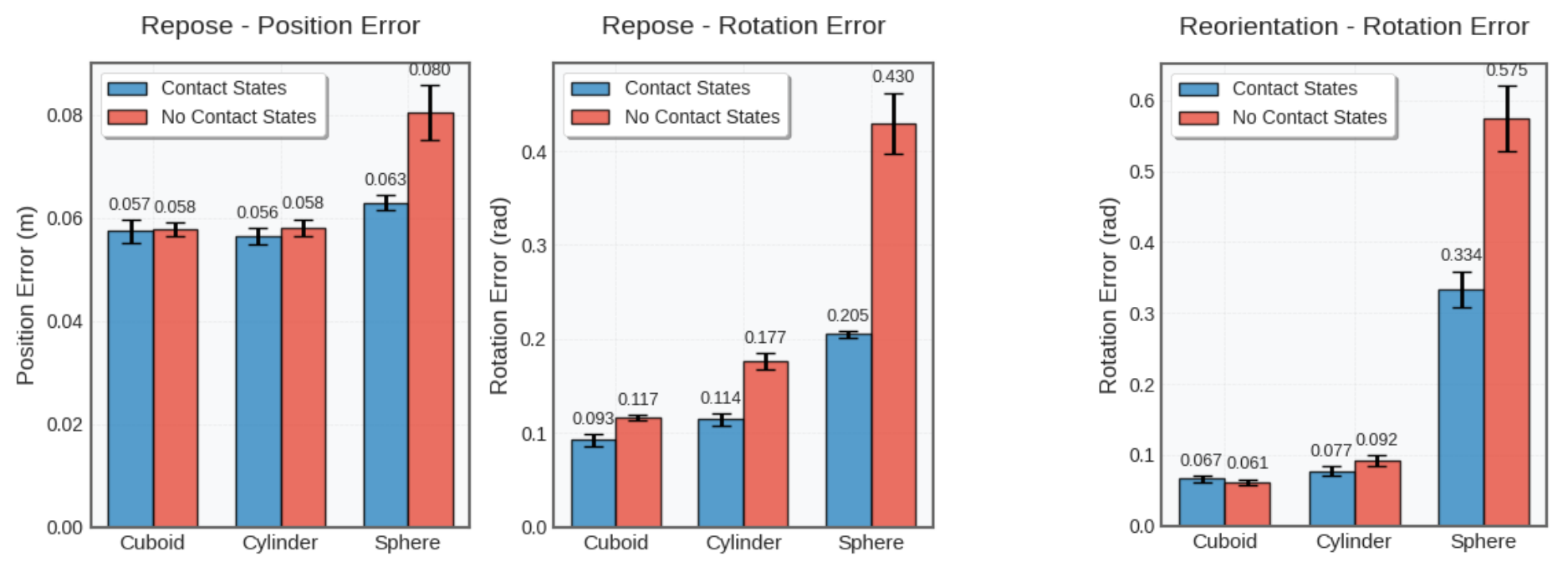} 
		\caption{Tracking error comparison on unseen object shapes across various tasks. We compare our \textcolor{blue}{contact-explicit policy} (uses contact states) with a baseline policy that uses \textcolor{red}{one-hot task encoding} (no contact states) to represent the task. Results are averaged over 1000 episodes, each lasting 20 seconds of simulated time.}
		\label{fig:ood_shape}
	\end{figure*}

    \begin{table}[h]
    \centering
    \renewcommand{\arraystretch}{1.2}
    \begin{tabular}{|l|c|c|}
        \hline
        \textbf{Task / Repose} & \textbf{Contact-Explicit} & \textbf{One-hot task} \\
        \hline
        Position Error (m) & $\textbf{0.115} \pm 0.003$ & $0.129 \pm 0.003$ \\
        Rotation Error (rad) & $\textbf{0.390} \pm 0.006$ & $0.455 \pm 0.007$ \\
        \hline
        \textbf{Task / Reorientation} & \textbf{Contact-Explicit} & \textbf{One-hot task} \\
        \hline
        Rotation Error (rad) & $\textbf{0.109} \pm 0.002$ & $0.191 \pm 0.004$ \\
        \hline
    \end{tabular}
    \caption{
    Quantitative evaluation of our contact-explicit policy against a one-hot task encoding baseline on out-of-distribution object poses. Object poses were sampled from $\mathcal{U}_{\text{train}}$ / $\mathcal{U}_{\text{eval}}$, where the first range corresponds to training and the second to evaluation:
        $X\!\sim\!\mathcal{U}_{\text{train}}(0.0, 0.1)$ / $\mathcal{U}_{\text{eval}}(0.0, 0.2)$,
        $Y\!\sim\!\mathcal{U}_{\text{train}}(-0.15, 0.15)$ / $\mathcal{U}_{\text{eval}}(-0.3, 0.3)$,
        $Z\!\sim\!\mathcal{U}_{\text{train}}(0.05, 0.25)$ / $\mathcal{U}_{\text{eval}}(0.05, 0.4)$,
        and roll, pitch, yaw $\!\sim\!\mathcal{U}_{\text{train}}(-0.6, 0.6)$ / $\mathcal{U}_{\text{eval}}(-1.2, 1.2)$.
    }
    \label{tab:pose_errors}
\end{table}

    \textbf{Manipulation generalization to unseen object shapes.} We compare our bimanual manipulation policy against a policy that instead uses one-hot task encoding to distinguish the tasks. Here, we evaluate the performance of the two policies, one with contact goals in the state (ours, blue) and another that uses a one-hot task encoding (baseline, red) instead to distinguish the tasks. For the baseline, we additionally provide the object dimensions as an observation to the policy. The policy was only trained on cuboidal shapes and evaluated here with cylindrical and spherical shapes. Our results are summarised in Fig.~\ref{fig:ood_shape}. 

    As with the contact-explicit locomotion policy being able to generalize to unseen contact locations (velocity directions), we witness with the contact-explicit bimanual manipulation policy that we can better generalize to unseen object shapes. Using the contact-explicit approach, the policy learns to track the contact locations on the different shapes much better than other implicit goal representations. Especially in the case of spherical shapes for object reorientation on the table, we observe that our policy comes up with emergent retrying behavior to track the object poses as the object starts rolling on the table.
    

    \textbf{Manipulation generalization to unseen object poses.} We compare the two bimanual manipulation policies mentioned in the previous experiment, contact-explicit (ours) and one-hot task encoding (baseline), to track object poses that were outside the training distribution. This was done by extrapolating the range of object poses from those seen in the training distribution.  The results, summarized in Table~\ref{tab:pose_errors}, demonstrate that contact-explicit policy consistently outperforms the one-hot task policy, with lower variability indicating more stable task accomplishment. This underscores the value of contact-aware policies for robust bimanual manipulation in uncertain scenarios.






\section{Conclusion}
\label{sec:conclusion}
We presented a unified contact-explicit task representation for learning a wide range of locomotion and manipulation skills through reinforcement learning. By treating contact as a central physical primitive, rather than a byproduct of motion optimization, our approach enables a single policy to generalize across morphologically diverse platforms and tasks. Empirical results demonstrate that contact-explicit policies offer stronger generalization to out-of-distribution goal configurations.

Moving forward, we aim to extend this framework to hierarchical reinforcement learning by coupling our low-level contact-conditioned policy with a learned high-level planner. This would allow autonomous long-horizon loco-manipulation in complex environments. We also plan to explore prehensile interactions and improve sim-to-real robustness for real-world deployment.


\clearpage

\acks{This work was partially supported by the Huawei-TUM joint laboratory.}

\bibliography{l4dc2026-sample}

@article{
	peng2017deeploco,
	author = {Peng, Xue Bin and Berseth, Glen and Yin, Kangkang and Van De Panne, Michiel},
	title = {DeepLoco: Dynamic Locomotion Skills Using Hierarchical Deep Reinforcement Learning},
	journal = {ACM Trans. Graph.},
	issue_date = {July 2017},
	volume = {36},
	number = {4},
	month = jul,
	year = {2017},
	issn = {0730-0301},
	pages = {41:1--41:13},
	articleno = {41},
	numpages = {13},
	url = {http://doi.acm.org/10.1145/3072959.3073602},
	doi = {10.1145/3072959.3073602},
	acmid = {3073602},
	publisher = {ACM},
	address = {New York, NY, USA},
}

@misc{zhang2024wococo,
      title={WoCoCo: Learning Whole-Body Humanoid Control with Sequential Contacts}, 
      author={Chong Zhang and Wenli Xiao and Tairan He and Guanya Shi},
      year={2024},
      eprint={2406.06005},
      archivePrefix={arXiv},
      primaryClass={cs.RO}
}

@article{lin2025sim,
          author={Lin, Toru and Sachdev, Kartik and Fan, Linxi and Malik, Jitendra and Zhu, Yuke},
          title={Sim-to-Real Reinforcement Learning for Vision-Based Dexterous Manipulation on Humanoids},
          journal={arXiv:2502.20396},
          year={2025}
        }

@article{margolis2022walktheseways,
    title={Walk These Ways: Tuning Robot Control for Generalization with Multiplicity of Behavior},
    author={Margolis, Gabriel B and Agrawal, Pulkit},
    journal={Conference on Robot Learning},
    year={2022}
}

@inproceedings{omar23humanoids,
  author = {Omar, Shafeef and Amatucci, Lorenzo and Turrisi, Giulio and Barasuol, Victor. and Semini, Claudio},
  title = {SafeSteps: Learning Safer Footstep Planning Policies for Legged Robots via Model-Based Priors},
  booktitle = {IEEE-RAS International Conference on Humanoid Robots},
  year = {2023},
}

@article{he2024hover,
      title={Hover: Versatile neural whole-body controller for humanoid robots},
      author={He, Tairan and Xiao, Wenli and Lin, Toru and Luo, Zhengyi and Xu, Zhenjia and Jiang, Zhenyu and Kautz, Jan and Liu, Changliu and Shi, Guanya and Wang, Xiaolong and others},
      journal={arXiv preprint arXiv:2410.21229},
      year={2024}
}

@misc{schulman2017proximalpolicyoptimizationalgorithms,
      title={Proximal Policy Optimization Algorithms}, 
      author={John Schulman and Filip Wolski and Prafulla Dhariwal and Alec Radford and Oleg Klimov},
      year={2017},
      eprint={1707.06347},
      archivePrefix={arXiv},
      primaryClass={cs.LG},
      url={https://arxiv.org/abs/1707.06347}, 
}

@misc{yin2025dexteritygenfoundationcontrollerunprecedented,
      title={DexterityGen: Foundation Controller for Unprecedented Dexterity}, 
      author={Zhao-Heng Yin and Changhao Wang and Luis Pineda and Francois Hogan and Krishna Bodduluri and Akash Sharma and Patrick Lancaster and Ishita Prasad and Mrinal Kalakrishnan and Jitendra Malik and Mike Lambeta and Tingfan Wu and Pieter Abbeel and Mustafa Mukadam},
      year={2025},
      eprint={2502.04307},
      archivePrefix={arXiv},
      primaryClass={cs.RO},
      url={https://arxiv.org/abs/2502.04307}, 
}

@misc{zhang2024learningagilelocomotionrisky,
      title={Learning Agile Locomotion on Risky Terrains}, 
      author={Chong Zhang and Nikita Rudin and David Hoeller and Marco Hutter},
      year={2024},
      eprint={2311.10484},
      archivePrefix={arXiv},
      primaryClass={cs.RO},
      url={https://arxiv.org/abs/2311.10484}, 
}

@misc{andrychowicz2018hindsightexperiencereplay,
      title={Hindsight Experience Replay}, 
      author={Marcin Andrychowicz and Filip Wolski and Alex Ray and Jonas Schneider and Rachel Fong and Peter Welinder and Bob McGrew and Josh Tobin and Pieter Abbeel and Wojciech Zaremba},
      year={2018},
      eprint={1707.01495},
      archivePrefix={arXiv},
      primaryClass={cs.LG},
      url={https://arxiv.org/abs/1707.01495}, 
}

@article{mittal2023orbit,
  title={Orbit: A Unified Simulation Framework for Interactive Robot Learning Environments}, 
  author={Mittal, Mayank and Yu, Calvin and Yu, Qinxi and Liu, Jingzhou and Rudin, Nikita and Hoeller, David and Yuan, Jia Lin and Singh, Ritvik and Guo, Yunrong and Mazhar, Hammad and Mandlekar, Ajay and Babich, Buck and State, Gavriel and Hutter, Marco and Garg, Animesh},
  journal={IEEE Robotics and Automation Letters}, 
  year={2023},
  volume={8},
  number={6},
  pages={3740-3747},
  doi={10.1109/LRA.2023.3270034}
}

@article{li2024reinforcement,
  title={Reinforcement learning for versatile, dynamic, and robust bipedal locomotion control},
  author={Li, Zhongyu and Peng, Xue Bin and Abbeel, Pieter and Levine, Sergey and Berseth, Glen and Sreenath, Koushil},
  journal={The International Journal of Robotics Research},
  pages={02783649241285161},
  year={2024},
  publisher={SAGE Publications Sage UK: London, England}
}

@article{ciebielski2024contact,
  title={Contact-conditioned learning of locomotion policies},
  author={Ciebielski, Michal and Khadiv, Majid},
  journal={arXiv preprint arXiv:2408.00776},
  year={2024}
}

@article{bellegarda2024allgaits,
  title={AllGaits: Learning All Quadruped Gaits and Transitions},
  author={Bellegarda, Guillaume and Shafiee, Milad and Ijspeert, Auke},
  journal={arXiv preprint arXiv:2411.04787},
  year={2024}
}

@article{chen2022towards,
  title={Towards human-level bimanual dexterous manipulation with reinforcement learning},
  author={Chen, Yuanpei and Wu, Tianhao and Wang, Shengjie and Feng, Xidong and Jiang, Jiechuan and Lu, Zongqing and McAleer, Stephen and Dong, Hao and Zhu, Song-Chun and Yang, Yaodong},
  journal={Advances in Neural Information Processing Systems},
  volume={35},
  pages={5150--5163},
  year={2022}
}

@article{lin2023bi,
  title={Bi-touch: Bimanual tactile manipulation with sim-to-real deep reinforcement learning},
  author={Lin, Yijiong and Church, Alex and Yang, Max and Li, Haoran and Lloyd, John and Zhang, Dandan and Lepora, Nathan F},
  journal={IEEE Robotics and Automation Letters},
  volume={8},
  number={9},
  pages={5472--5479},
  year={2023},
  publisher={IEEE}
}

@article{pan2025roboduet,
  title={RoboDuet: Learning a Cooperative Policy for Whole-body Legged Loco-Manipulation},
  author={Pan, Guoping and Ben, Qingwei and Yuan, Zhecheng and Jiang, Guangqi and Ji, Yandong and Li, Shoujie and Pang, Jiangmiao and Liu, Houde and Xu, Huazhe},
  journal={IEEE Robotics and Automation Letters},
  year={2025},
  publisher={IEEE}
}

@inproceedings{dao2024sim,
  title={Sim-to-real learning for humanoid box loco-manipulation},
  author={Dao, Jeremy and Duan, Helei and Fern, Alan},
  booktitle={2024 IEEE International Conference on Robotics and Automation (ICRA)},
  pages={16930--16936},
  year={2024},
  organization={IEEE}
}

@article{liu2024visual,
  title={Visual whole-body control for legged loco-manipulation},
  author={Liu, Minghuan and Chen, Zixuan and Cheng, Xuxin and Ji, Yandong and Qiu, Ri-Zhao and Yang, Ruihan and Wang, Xiaolong},
  journal={arXiv preprint arXiv:2403.16967},
  year={2024}
}

@article{qiu2024wildlma,
  title={WildLMa: Long Horizon Loco-Manipulation in the Wild},
  author={Qiu, Ri-Zhao and Song, Yuchen and Peng, Xuanbin and Suryadevara, Sai Aneesh and Yang, Ge and Liu, Minghuan and Ji, Mazeyu and Jia, Chengzhe and Yang, Ruihan and Zou, Xueyan and others},
  journal={arXiv preprint arXiv:2411.15131},
  year={2024}
}

@inproceedings{sleiman2024guided,
  title={Guided Reinforcement Learning for Robust Multi-Contact Loco-Manipulation},
  author={Sleiman, Jean-Pierre and Mittal, Mayank and Hutter, Marco},
  booktitle={8th Annual Conference on Robot Learning (CoRL 2024)},
  year={2024}
}

@article{tan2018sim,
  title={Sim-to-real: Learning agile locomotion for quadruped robots},
  author={Tan, Jie and Zhang, Tingnan and Coumans, Erwin and Iscen, Atil and Bai, Yunfei and Hafner, Danijar and Bohez, Steven and Vanhoucke, Vincent},
  journal={arXiv preprint arXiv:1804.10332},
  year={2018}
}

@article{hwangbo2019learning,
  title={Learning agile and dynamic motor skills for legged robots},
  author={Hwangbo, Jemin and Lee, Joonho and Dosovitskiy, Alexey and Bellicoso, Dario and Tsounis, Vassilios and Koltun, Vladlen and Hutter, Marco},
  journal={Science Robotics},
  volume={4},
  number={26},
  pages={eaau5872},
  year={2019},
  publisher={American Association for the Advancement of Science}
}

@misc{sferrazza2024humanoidbench,
    title={HumanoidBench: Simulated Humanoid Benchmark for Whole-Body Locomotion and Manipulation},
    author={Carmelo Sferrazza and Dun-Ming Huang and Xingyu Lin and Youngwoon Lee and Pieter Abbeel},
    year={2024},
}

@misc{rudin2022advancedskillslearninglocomotion,
      title={Advanced Skills by Learning Locomotion and Local Navigation End-to-End}, 
      author={Nikita Rudin and David Hoeller and Marko Bjelonic and Marco Hutter},
      year={2022},
      eprint={2209.12827},
      archivePrefix={arXiv},
      primaryClass={cs.RO},
      url={https://arxiv.org/abs/2209.12827}, 
}

@misc{hoeller2023anymalparkourlearningagile,
      title={ANYmal Parkour: Learning Agile Navigation for Quadrupedal Robots}, 
      author={David Hoeller and Nikita Rudin and Dhionis Sako and Marco Hutter},
      year={2023},
      eprint={2306.14874},
      archivePrefix={arXiv},
      primaryClass={cs.RO},
      url={https://arxiv.org/abs/2306.14874}, 
}

@article{cheng2023parkour,
title={Extreme Parkour with Legged Robots},
author={Cheng, Xuxin and Shi, Kexin and Agarwal, Ananye and Pathak, Deepak},
journal={arXiv preprint arXiv:2309.14341},
year={2023}
}

@misc{rudin2022learningwalkminutesusing,
      title={Learning to Walk in Minutes Using Massively Parallel Deep Reinforcement Learning}, 
      author={Nikita Rudin and David Hoeller and Philipp Reist and Marco Hutter},
      year={2022},
      eprint={2109.11978},
      archivePrefix={arXiv},
      primaryClass={cs.RO},
      url={https://arxiv.org/abs/2109.11978}, 
}

@misc{openai2019solvingrubikscuberobot,
      title={Solving Rubik's Cube with a Robot Hand}, 
      author={OpenAI and Ilge Akkaya and Marcin Andrychowicz and Maciek Chociej and Mateusz Litwin and Bob McGrew and Arthur Petron and Alex Paino and Matthias Plappert and Glenn Powell and Raphael Ribas and Jonas Schneider and Nikolas Tezak and Jerry Tworek and Peter Welinder and Lilian Weng and Qiming Yuan and Wojciech Zaremba and Lei Zhang},
      year={2019},
      eprint={1910.07113},
      archivePrefix={arXiv},
      primaryClass={cs.LG},
      url={https://arxiv.org/abs/1910.07113}, 
}

@misc{singh2025dextrahrgbvisuomotorpoliciesgrasp,
      title={DextrAH-RGB: Visuomotor Policies to Grasp Anything with Dexterous Hands}, 
      author={Ritvik Singh and Arthur Allshire and Ankur Handa and Nathan Ratliff and Karl Van Wyk},
      year={2025},
      eprint={2412.01791},
      archivePrefix={arXiv},
      primaryClass={cs.RO},
      url={https://arxiv.org/abs/2412.01791}, 
}

@misc{fu2022deepwholebodycontrollearning,
      title={Deep Whole-Body Control: Learning a Unified Policy for Manipulation and Locomotion}, 
      author={Zipeng Fu and Xuxin Cheng and Deepak Pathak},
      year={2022},
      eprint={2210.10044},
      archivePrefix={arXiv},
      primaryClass={cs.RO},
      url={https://arxiv.org/abs/2210.10044}, 
}

@misc{taouil2025physicallyconsistenthumanoidlocomanipulation,
      title={Physically Consistent Humanoid Loco-Manipulation using Latent Diffusion Models}, 
      author={Ilyass Taouil and Haizhou Zhao and Angela Dai and Majid Khadiv},
      year={2025},
      eprint={2504.16843},
      archivePrefix={arXiv},
      primaryClass={cs.RO},
      url={https://arxiv.org/abs/2504.16843}, 
}

@inproceedings{dhedin2024diffusion,
  title={Diffusion-based learning of contact plans for agile locomotion},
  author={Dhedin, Victor and Ravi, Adithya Kumar Chinnakkonda and Jordana, Armand and Zhu, Huaijiang and Meduri, Avadesh and Righetti, Ludovic and Sch{\"o}lkopf, Bernhard and Khadiv, Majid and others},
  booktitle={2024 IEEE-RAS 23rd International Conference on Humanoid Robots (Humanoids)},
  pages={637--644},
  year={2024},
  organization={IEEE}
}

@misc{zargarbashi2024robotkeyframinglearninglocomotionhighlevel,
      title={RobotKeyframing: Learning Locomotion with High-Level Objectives via Mixture of Dense and Sparse Rewards}, 
      author={Fatemeh Zargarbashi and Jin Cheng and Dongho Kang and Robert Sumner and Stelian Coros},
      year={2024},
      eprint={2407.11562},
      archivePrefix={arXiv},
      primaryClass={cs.RO},
      url={https://arxiv.org/abs/2407.11562}, 
}

@misc{lum2024dextrahgpixelstoactiondexterousarmhand,
      title={DextrAH-G: Pixels-to-Action Dexterous Arm-Hand Grasping with Geometric Fabrics}, 
      author={Tyler Ga Wei Lum and Martin Matak and Viktor Makoviychuk and Ankur Handa and Arthur Allshire and Tucker Hermans and Nathan D. Ratliff and Karl Van Wyk},
      year={2024},
      eprint={2407.02274},
      archivePrefix={arXiv},
      primaryClass={cs.RO},
      url={https://arxiv.org/abs/2407.02274}, 
}

@article{
miki2022perceptivequadrupedallocomotion,
author = {Takahiro Miki  and Joonho Lee  and Jemin Hwangbo  and Lorenz Wellhausen  and Vladlen Koltun  and Marco Hutter },
title = {Learning robust perceptive locomotion for quadrupedal robots in the wild},
journal = {Science Robotics},
volume = {7},
number = {62},
pages = {eabk2822},
year = {2022},
doi = {10.1126/scirobotics.abk2822},
URL = {https://www.science.org/doi/abs/10.1126/scirobotics.abk2822},
eprint = {https://www.science.org/doi/pdf/10.1126/scirobotics.abk2822}}

\end{document}